\pdfoutput=1

\documentclass[11pt]{article}

\usepackage[hyperref]{emnlp2021}

\usepackage{times} 
\usepackage{latexsym}

\usepackage[T1]{fontenc}

\usepackage[utf8]{inputenc}

\usepackage{microtype}

\usepackage{graphicx}
\usepackage{caption}
\usepackage{subcaption}
\usepackage{booktabs}
\usepackage{multirow}
\usepackage{multicol}
\usepackage{amsmath}
\usepackage{courier}
\usepackage{url}
\usepackage{CJKutf8}

\usepackage{microtype}
\usepackage{textcomp}

\usepackage{enumitem}
\setlist{nosep} 

\usepackage{tikz}
\usepackage{pgfplots}
\pgfplotsset{width=7.5cm,compat=1.12}
\usepgfplotslibrary{fillbetween}

\usepackage{makecell}

\usepackage{xcolor}

\newcommand*{\dittoclosing}{\raisebox{-0.5ex}{''} }

\title{MeetDot: Videoconferencing with Live Translation Captions}

\author{Arkady Arkhangorodsky, Christopher Chu, Scot Fang, Yiqi Huang, \\ \bf Denglin Jiang, Ajay Nagesh, Boliang Zhang, Kevin Knight  \\
DiDi Labs\\
4640 Admiralty Way \\
Marina del Rey, CA 90292 \\
{\small \tt \{arkadyarkhangorodsky,chrischu,scotfang,denglinjiang,} \\ {\small \tt yiqihuang,ajaynagesh,boliangzhang,kevinknight\}@didiglobal.com}}

\begin{document}

\maketitle

\begin{abstract}
We present {\em MeetDot}, a videoconferencing system with live translation captions overlaid on screen. The system aims to facilitate conversation between people who speak different languages, thereby reducing communication barriers between multilingual participants. Currently, our system supports speech and captions in 4 languages and combines automatic speech recognition (ASR) and machine translation (MT) in a cascade. We use the {\em re-translation} strategy to translate the streamed speech, resulting in caption flicker. Additionally, our system has very strict latency requirements to have acceptable call quality. We implement several features to enhance user experience and reduce their cognitive load, such as smooth scrolling captions and reducing caption flicker. The modular architecture allows us to integrate different ASR and MT services in our backend. Our system provides an integrated evaluation suite to optimize key intrinsic evaluation metrics such as accuracy, latency and erasure. Finally, we present an innovative cross-lingual word-guessing game as an extrinsic evaluation metric to measure end-to-end system performance. We plan to make our system open-source for research purposes.\footnote{The system will be available at \url{https://github.com/didi/meetdot}}
\end{abstract}

\section{Introduction}

As collaborations across countries is the norm in the modern workplace, videoconferencing is an indispensable part of our working lives. The recent widespread adoption of remote work has necessitated effective online communication tools, especially among people who speak different languages. In this work, we present {\em MeetDot}, a videoconferencing solution with live translation captions. Participants can see an overlaid translation of other participants' speech in their preferred language. Currently, we support speech and captions in English, Chinese, Spanish and Portuguese. 

The system is built by cascading automatic speech recognition (ASR) and machine translation (MT) components. We process the incoming speech signal in a streaming mode, transcribe it in the speaker's language to be used as input to an MT system to decode in the listener's language during the call. We have very tight latency requirements to be able to provide good quality captions in a live video call.  Our framework has several features to enable a better user experience and reduce the cognitive load on the participants such as smooth pixel-wise scrolling of the captions, fading text that is likely to change and biased decoding of machine translation output~\cite{bias-decoding} to reduce the flicker of the captions. In addition, we present preliminary work towards identifying named-entity mentions in speech by interpolating a pronunciation dictionary with the ASR language model.

We present the key metrics to measure the quality of the captions such as accuracy, latency and stability (flicker in captions). Our system provides an integrated evaluation suite to enable fast development through hill climbing on these metrics. In addition to these intrinsic metrics, we present an interesting online cross-lingual word guessing game, 
where one of the participants is given a word that they describe and the other participants have to guess the word by reading the captions in their respective languages. 

\begin{figure*}[hbt!]
    \centering
    \includegraphics[width=5in]{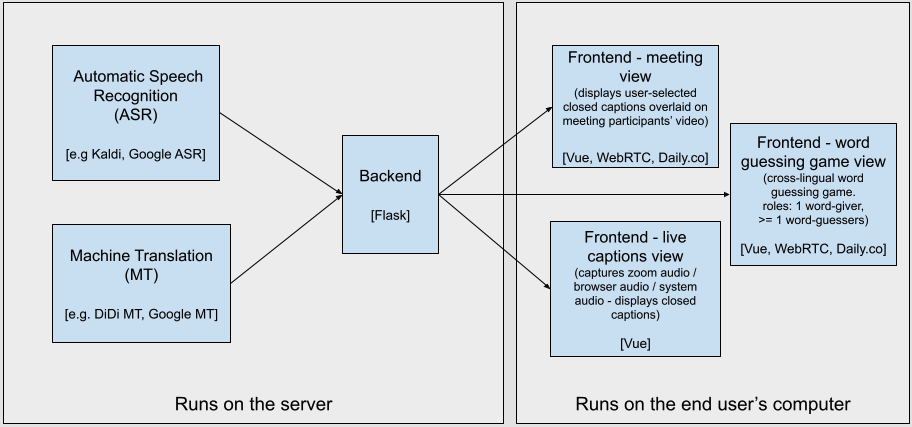}
    \caption{MeetDot system architecture. The ASR and MT modules are designed to be plug-and-play with different services in the backend (e.g. in-house DiDi MT system/Google MT API). Frontend consists of different ``views'' - e.g. live captions view - captures audio from another browser window/zoom/system to show translated captions. }
    \label{fig:sysarch}
\end{figure*}

We present the following as the key contributions of our work: 
\begin{itemize}
    \item A video conference system with live translation of multilingual speech into captions overlaid on participants' videos. The system has several features to enhance user experience.
    \item A comprehensive evaluation suite closely integrated with the system and a set of metrics to reduce latency, caption flicker and accuracy. 
    \item A cross-lingual word-guessing game that can be used as an extrinsic metric to evaluate end-to-end system performance. 
\end{itemize}

We are in the process of releasing the system as open-source software for the purpose of furthering research in the area.

\section{Related Work}

The area of simultaneous translation has attracted a lot of attention is recent years. The recent editions of the Workshop on Automatic Simultaneous Translation~\citep{autosimtrans-2021,autosimtrans-2020} and the tutorial in EMNLP 2020~\cite{tutorial_emnlp20} provide us an overview of the state-of-the-art and challenges in live translation. Recent technical advances include newer architectures such as prefix-to-prefix~\cite{prefix2prefix} and adaptive policy methods such as imitation learning~\cite{imitation} and monotonic attention~\cite{monotonic_attention}. The solutions in this space are differentiated into speech-to-speech and speech-to-text, in the former there is a speech synthesis component. Our work falls in the latter bucket, since we only display captions to the user. Since we have access to separate audio-input channels (one per participant), we do not need speaker diarization~\cite{diarization}. 

{\em Re-translation} is a common strategy applied, wherein we translate from scratch every new extended source sentence, transcribed through the ASR module as the participant speaks. The ability to modify previously displayed captions when new ASR output is available can lead to flicker in displayed captions. Following previous work~\cite{bias-decoding}, we measure this using the erasure metric and reduce it using biased beam search during MT decoding. To better capture our use case of short exchanges in a meeting scenario compared to long speech, we introduce additional metrics {\em initial lag}, {\em incremental caption lag}, {\em mean word burstiness} and {\em max word burstiness}. 

There have been recent working systems in this space, such as~\citet{prefix2prefix,german_live_translation,ted_live_translation} that provide live captions for single-speaker lectures and does not focus on multi-party meetings. Very recently, there are news reports of systems that offer live translations of multi-party meetings~\cite{CiscoWebex}, but their technical details are unclear. 


\section{System Description}
\label{sec:sys_desc}

\begin{figure*}[hbt!]
    \centering
    \includegraphics[width=5.2in]{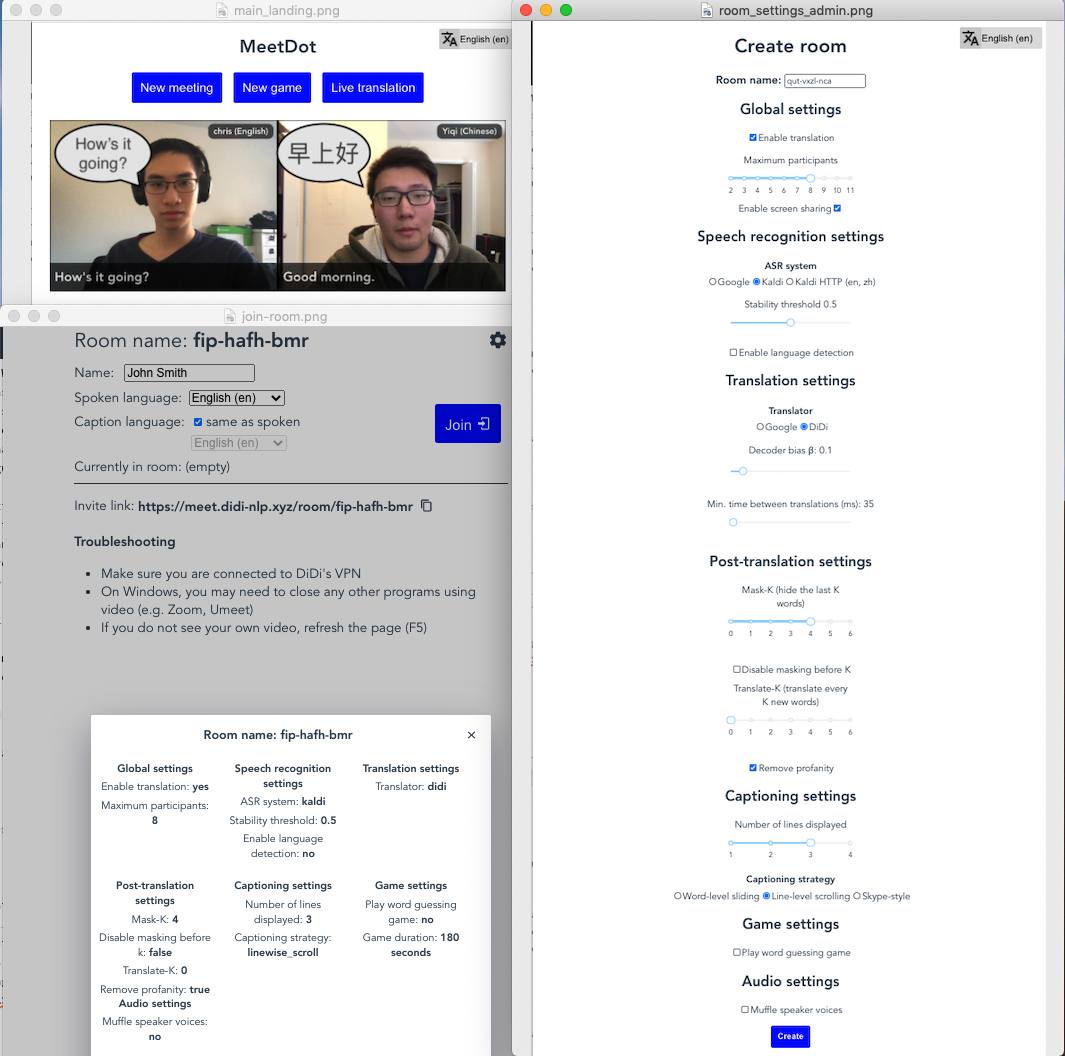}
    \caption{MeetDot room creation. Landing page (top, left panel) Any user can set up a MeetDot room and share its URL with potential participants (bottom, left panel).  Admin users can select parameters that control captioning, speech recognition, and translation (right panel, \textsection \ref{sec:sys_desc}).}
    \label{fig:create-room-ui}
\end{figure*}

The overall system architecture is shown in Figure~\ref{fig:sysarch}. It mainly consists of two components (1) {\em Frontend}: runs on the user's computer locally (2) {\em Backend}: runs on the server and consists of ASR and MT modules that interact with the Flask server. The modular architecture allows us to swap the ASR and MT components from different service providers (such as Google ASR or in-house Kaldi-based ASR services). These are explained below. \\

\begin{figure*}
    \centering
    \includegraphics[width=6in]{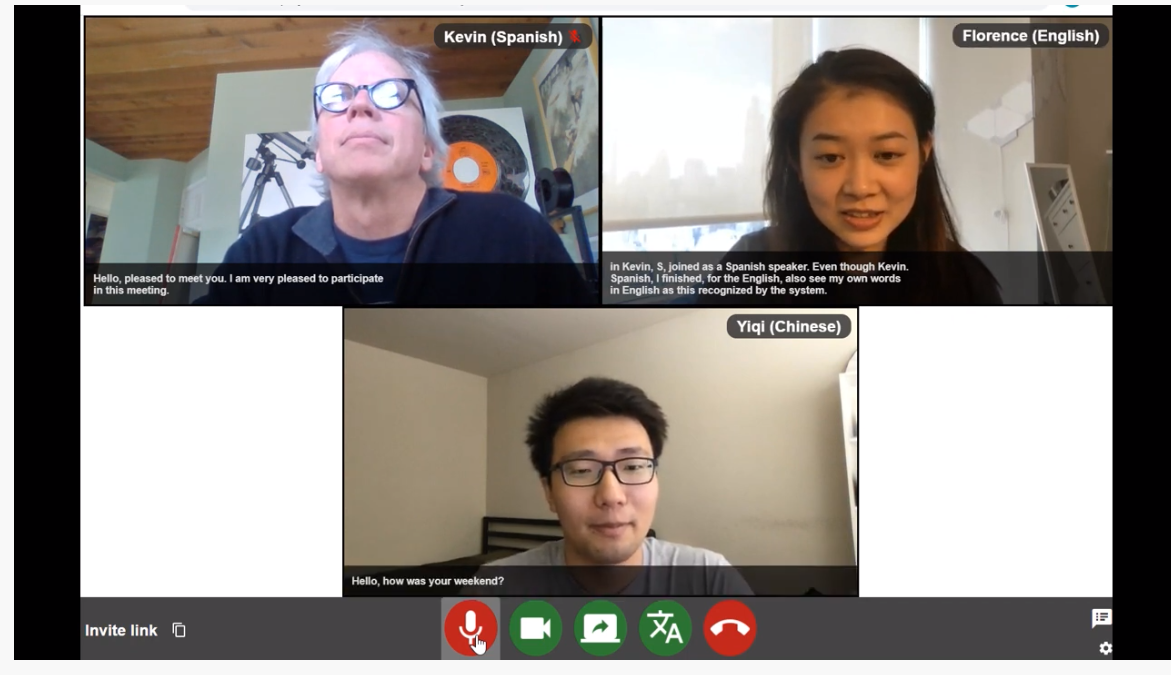}
    \caption{MeetDot videoconference interface. Translated captions are incrementally updated (word-by-word, phrase-by-phrase) on top of participant videos.  Translations also appear in the transcript panel (on right, not shown), updated utterance-by-utterance. Choosing a caption language (4th button from left at the bottom, in green) displays all captions in that particular language. This depicts the view of the English caption user.}
    \label{fig:meetdot-ui}
\end{figure*}

\noindent\textbf{MeetDot User Interface: }
We implemented a simple web-based user interface that allows users to have meetings with automatically translated captions overlaid. Our interface consists of: \\
\noindent - A home page (Figure \ref{fig:create-room-ui}, top-left panel) for creating a meeting room with default settings, or a game (\textsection~\ref{sec:eval}).\\
\noindent - A meeting creation page, where the room settings can be configured in different ways for experimentation. (Figure \ref{fig:create-room-ui}, right panel)\\
\noindent - A meeting landing page, where a user specifies their speaking language and caption language (usually the same) before joining the meeting. (Figure \ref{fig:create-room-ui}, bottom-left panel)\\
\noindent - A meeting page, where the video call takes place (Figure \ref{fig:meetdot-ui})

The meeting page features real-time translated captions overlaid on each speaker's video, displayed to each user in their selected language (Figure \ref{fig:meetdot-ui}). The user's spoken language and caption language can also be changed on-the-fly. The meetings support full video-conferencing functionality, including microphone and camera toggles, screen sharing, and gallery/focused views. Finally, there is a panel that shows the automatically transcribed and translated conversation history in the user's preferred language, which can also be saved.

MeetDot also supports ``Live Translation'' mode that can translate audio from any audio feed, e.g. a microphone or the user's own computer. We provide instructions for the latter, so users can see captions for audio in either another browser tab or an ongoing Zoom call or audio from their system.

The frontend is implemented in Vue,\footnote{\url{https://vuejs.org/}} and the backend is implemented using Flask. Video-conferencing functionality is built with WebRTC,\footnote{\url{https://webrtc.org/}} using {\em Daily.co}\footnote{\url{https://www.daily.co/}} as the signaling server to initiate real-time audio and video connections. The frontend sends audio bytes to the Flask backend through a WebSocket connection, which routes them to the speech recognition and translation services and returns translated captions to be displayed.\\

\noindent\textbf{Speech Recognition and Machine Translation: }
Our backend services of ASR and MT are joined in a cascaded manner. Each participant's speech is fed in a streaming fashion to the ASR module of the appropriate language selected by the user. Additionally, we show the ASR output as captions to the speaker as feedback/confirmation to them. Each of the  transcribed text returned by ASR is fed to the MT system to translate it into the caption language selected by the reader, from scratch. This strategy is termed as {\em re-translation}. Since the ASR stream continually returns a revised or an extended string, the input to MT is noisy and will lead to the captions overwritten frequently (termed as {\em flicker}) leading to a high cognitive load on the reading. We employ several techniques to have better user experience while they are reading the captions (elaborated more below). 

At present, we support English, Chinese, Spanish and Portuguese languages for both speech and captions. The modular architecture of our system allows us to plug-and-play different ASR and MT service components in the backend. We develop two different ASR systems based on the Kaldi framework and WeNet~\cite{wenet}. We can also swap either of these to use Google ASR API instead. For MT, we have the option of using our in-house DiDi MT system~\cite{didi_mt} as well as call Google MT API. 
For Kaldi ASR, we adapt pre-trained Kaldi models to videoconferencing domain by interpolating the pre-trained language model with our in-domain language models. For WeNet, we use use the {\em Unified Conformer} model and language model interpolation is planned for future work.\footnote{We use a pre-trained checkpoint for English ASR and trained a Chinese ASR model from scratch using the multi-cn and TAL datasets.}  Following~\citet{bias-decoding}, we modify the decoding procedure of MT in OpenNMT's ctranslate toolkit\footnote{\url{https://github.com/OpenNMT/CTranslate2}} to mitigate the issue of flicker mentioned above. 

{
\begin{table*}[hbt!]
\scriptsize{
    \centering
    \begin{tabular}{|c|c||r|r|r||r|r|r|r|}
    \hline
Direction & Systems & {\bf Final} & {\bf Translation} & {\bf Normalized} & {\bf Initial} & {\bf Incremental} & {\bf Mean word} & {\bf Max word}\\
& & {\bf bleu} & {\bf lag (s)} & {\bf erasure} & {\bf lag (s)} & {\bf caption lag (s)} & {\bf burstiness} & {\bf burstiness} \\
    \hline
    \hline
En-to-Zh & {\em  Google ASR, Google MT} & 17.81 & 5.84 & 2.69 & 3.82 & 0.71 & 5.26 & 11.83 \\
\hline
\dittoclosing & {\em Kaldi ASR, DiDi MT} & 19.59 & 2.72 & 0.33 & 2.59 & 0.43 & 4.34 & 9.20 \\
\hline
\dittoclosing & {\em WeNet ASR, DiDi MT} & 23.76 & 2.39 & 0.21 & 2.73 & 0.47 & 4.76 & 9.62 \\
\hline
\hline
Zh-to-En & {\em Google ASR, Google MT}  & 9.99 & 2.33 & 0.47 & 4.31 & 2.12 & 5.20 & 9.46 \\
\hline
\dittoclosing & {\em Kaldi ASR, DiDi MT} & 7.88 & 3.33 & 0.73 & 2.70 & 0.42 & 2.42 & 6.13 \\
\hline
\dittoclosing & {\em WeNet ASR, DiDi MT} & 9.76 & 2.27 & 0.37 & 2.52 & 0.32 & 2.42 & 5.65 \\
\hline
\end{tabular}
    \caption{Baseline results on our {\em daily work conversation} dataset. Note that the Google API does not have biased-decoding available to reduce flicker. Metrics are explained in \textsection\ref{sec:eval}. Right 4 metrics are introduced by our work.}
    \label{tab:baseline_results}
}
\end{table*}
}
We include several additional features to enhance user experience. We use NVIDIA NeMo toolkit\footnote{\url{https://github.com/NVIDIA/NeMo}} to punctuate the captions and predict if the word should be capitalized or not, which makes the captions more readable. 
We have an initial named entity recognition module, to recognize mentions of participants' names in speech when using the Kaldi ASR system. This is performed by interpolating the ASR's language model with a language model trained on a synthesised corpus that contains participants' names. The name pronunciation dictionary required by Kaldi ASR is generated automatically by rules. 
Profanity is detected using a standard list of keywords and starred in both ASR and translation output. We have an experimental module to detect the speaker's language automatically (instead of being manually set by the user). The advantage of such a feature is to allow code-switching between multiple languages which is a common behavior among multilingual speakers~\cite{code_switch,code_switch_2}.\\


\begin{figure*}
    \centering
    \includegraphics[width=5in]{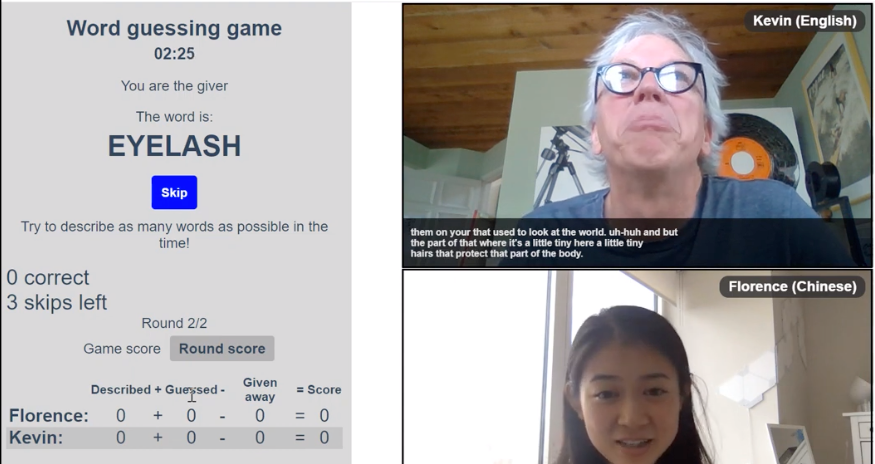}
    \caption{Cross-lingual word guessing game for extrinsic evaluation. Roles: one player is given a word to describe in their language, one or more players look at the captions displayed to guess the correct word. Screenshot shown is that of the describer of the word. Word is {\em ``eyelash''} (left panel) - the participants communicate hints and guesses through {\em MeetDot} translation captions, and the system itself spots correct guesses.} 
    \label{taboo}
\end{figure*}

\noindent\textbf{Captioning Strategies: }
Here we describe how we deploy ASR and MT capabilities to create translation captions that are incrementally updated in real time.  Since we display translation captions on top of a speaker's video feed, we have limited screen real-estate---for example, 3 lines of 60 characters each.

Each time we receive an incremental ASR update (hypothesis extension or revision), we translate the updated string from scratch \cite{bias-decoding}.  If the translation exceeds the available real-estate, we display the longest suffix that fits. ASR also signals utterance boundaries; we only send the current, growing utterance through MT, caching the MT results on previous utterances. 

The basic system exhibits large amounts of flicker \cite{flicker1, flicker2}, which we mitigate with these methods:

\noindent{\em Translate-k.} We only send every $k$th ASR output to MT. This results in captions that are more stable, but which update more slowly.

\noindent{\em Translate-t.} Improving on {translate-k}, we send an ASR output to MT if at least $t$ seconds have elapsed since the last MT call.

\noindent{\em Mask-k.} We suppress the last $k$ words from the MT output, providing time for the translation to settle down \cite{waitk2,waitk1,prefix2prefix,bias-decoding}.  We often use \texttt{Mask-4} in practice.

\noindent{\em Biased MT decoding.} We encourage word-by-word MT decoding to match the string output by the previous MT call \cite{bias-decoding}, avoiding translation variations that, while acceptable, unfortunately introduce flicker.

\noindent{\em Preserve linebreaks.} When possible, we prevent words from jumping back \& forth across linebreaks.

\noindent{\em Smooth scrolling.} We reduce perception of flicker by scrolling lines smoothly, pixel-wise.

\section{Evaluation}
\label{sec:eval}

\noindent\textbf{Dataset:} In order to evaluate our system in the most-appropriate deployment scenario, we construct an evaluation dataset based on meetings within our team. We have a total of 5 meetings, 3 meetings involving 7 participants and in English and 2 meetings involving 2 participants in Chinese. The content of the meetings are daily work conversation 
and contains a total of 135 mins (94 mins English and 41 mins Chinese). We manually transcribe and translate these meetings using a simple and uniform set of guidelines. 
The resulting dataset consists of 494 English utterances ($\sim11k$ words, translated into Chinese) and 183 Chinese utterances ($\sim9.8k$ characters, translated into English). We use this dataset to measure our intrinsic evaluation metrics.\\



\noindent\textbf{Intrinsic metrics: } We adopt the following metrics from previous work on streaming translation~\cite{bleu,flicker1,bias-decoding}:

\noindent - {\em Final Bleu}.  We measure the Bleu MT accuracy of final, target-language utterances against a reference set of human translations. Anti-flicker devices that ``lock in'' partial translations will generally decrease final Bleu.\\
\noindent - {\em Translation lag}.  We measure (roughly) the average difference, in seconds, between when a word was spoken and when its translation was finalized on the screen.\\
\noindent - {\em Normalized erasure}.  We quantify flicker as $m$/$n$, the number of words $m$ that get erased during the production of an $n$-word utterance.

To support our videoconferencing application, we introduce other intrinsic metrics:

\noindent - {\em Initial lag}.  In videoconferencing, we find it valuable to generate translation captions immediately after a speaker begins, even if the initial translation involves some flicker. Here we measure the time between initial speaking time and the first displayed word.  We improve initial lag by adopting a \texttt{ Mask-0} policy at the start of an utterance, transitioning to our usual \texttt{Mask-4} policy later.\\
\noindent - {\em Incremental caption lag}.  The average time between caption updates. \\
\noindent - {\em Mean (\& Max) word burstiness}. The mean number of words or characters for Chinese (maximum, respectively) that are added/subtracted in a single caption update (for a given utterance's translation--averaged over all utterances, respectively). 
%

In Table~\ref{tab:baseline_results}, we present baseline results of the various intrinsic metrics for the English to Chinese and Chinese to English systems. We present results for 3 different module combinations, namely, (i) Google API\footnote{Accessed on 2021-09-09.} for ASR\footnote{\url{https://cloud.google.com/speech-to-text}} and MT\footnote{\url{https://cloud.google.com/translate}} (ii) Kaldi ASR and DiDi MT (iii) WeNet ASR and DiDi MT. From the results table, we would like to highlight that the biased decoding modification for machine translation has a positive impact on several metrics such as translation lag, normalized erasure and word burstiness. Biased decoding is absent in the Google MT API, hence has higher numbers in these metrics. This  results in increasing flicker leading to a poorer user experience. Our final bleu score when using WeNet ASR and DiDi MT is several points better than Google ASR and Google MT in the English to Chinese direction and has comparable performance in Chinese to English direction. ASR system's performance is an important factor in a cascaded system. Kaldi ASR has a word error rate (WER) rate of 43.88 for English (character error rate (CER) of 49.75 for Chinese, respectively) compared to WeNet ASR's WER of 34.74 (CER of 38.03 for Chinese, respectively). This has a direct impact on final bleu scores as seen from the results. \\

\noindent\textbf{Cross-lingual word guessing game: } Extrinsic metrics for speech translation are not as popular as intrinsic ones~\cite{bias-decoding,flicker1}. However, given the broad range of techniques for displaying translation captions, we would like to measure things that are closer to the user experience.

Here, we introduce a cross-lingual, cooperative word game for A/B testing different captioning algorithms.  Players who speak different languages use {\em MeetDot}'s translation captions to communicate with each other. If the players obtain higher scores under one condition, we conclude that their communication is, in some way, made more effective and efficient.

The game is a variation on {\em Taboo},\footnote{\url{https://www.hasbro.com/common/instruct/Taboo(2000).PDF}} in which one player receives a secret word (such as ``racoon'' or ``scary'') and must describe it without mentioning the word or variant of it. The other player tries to guess the word.  The players are awarded a point for every word guessed in a 4-minute period. The first player is allowed to skip upto three words.

In our variation, the first player may receive a Chinese word and describe it in Chinese, while the second player sees English captions and makes guesses in English. When translation is quick, accurate, and readable, players score higher.

We design our bilingual wordlists to contain words with limited ambiguity. This way, we are able to build a reliable, automatic scorer that rewards players and advances them to the next word.

We also implement a competitive, multi-player version of the game, where players are assigned points for making correct guesses faster than others, and for giving clues that lead to fast guesses.

\section{Conclusion and Future Work}

We describe {\em MeetDot}, a videoconferencing system with live translation captions, along with its components: UI, ASR, MT, and captioning. We implement an evaluation suite that allows us to accurately compute metrics from the sequence of captions that users would see. We also describe a cross-lingual word game for A/B testing different captioning algorithms and conditions.

Our future work includes improved ASR/MT, extrinsic testing, and an open source release. Our overall goal is to provide a platform for developing translation captions that are accurate and ``right behind you''.

\bibliography{acl2020,dialog}
\bibliographystyle{acl_natbib}

\end{document}